\documentclass[nohyperref]{article}

\usepackage{microtype}
\usepackage{graphicx}
\usepackage{subfigure}
\usepackage{booktabs} 
\usepackage{hyperref}

\usepackage[ruled,vlined,linesnumbered,commentsnumbered]{algorithm2e}
  \SetKwInput{KwIn}{Inputs}
  \SetKwFunction{Fselfrepair}{SR}
  \SetKwFunction{Frepair}{SR-Layer}
  \SetKwFunction{Fsolve}{FindSatConstraint}
  \SetKwFunction{Freorder}{Reorder}
  \SetKwFunction{Ftopsort}{TopologicalSort}
  \SetKwFunction{Fcheckcycle}{ContainsCycle}
  \SetKwFunction{Ffilter}{Filter}
  \SetKwFunction{Ftodnf}{ToDNF}
  \SetKwFunction{Fprioritize}{Prioritize}
  \SetKwFunction{Fordergraph}{OrderGraph}
  \SetKwFunction{Fcontainscycle}{ContainsCycle}
  \SetKwFunction{Fissat}{IsSat}
  \SetKwFunction{Freorder}{Repair}
  \SetKwFunction{Fdescsort}{SortDescending}
  \SetKwFunction{Finvperm}{InvertPermutation}
  \SetKwProg{Fn}{}{:}{}
  \DontPrintSemicolon
  \newlength{\commentWidth}
\usepackage{tikz}
\usepackage{graphicx}
\usepackage{nicefrac}
\usepackage{multicol}
\usepackage{amsthm}

\usepackage{dsfont}
\usetikzlibrary{calc,decorations.markings}

\newcommand{\layername}{\ensuremath{\mathsf{SMTLayer}}\xspace}

\newcommand{\forward}{\ensuremath{\mathsf{F}_{\mathit{smt}}}\xspace}
\newcommand{\forwardmax}{\ensuremath{\mathsf{F}_{\mathit{max}}}\xspace}
\newcommand{\backcore}{\ensuremath{\mathsf{B}_{\mathit{core}}}\xspace}
\newcommand{\backmax}{\ensuremath{\mathsf{B}_{\mathit{max}}}\xspace}

\newcommand{\T}{\ensuremath{T}\xspace}

\newcommand{\X}{\ensuremath{\mathbf{X}}\xspace}
\newcommand{\Y}{\ensuremath{\mathbf{Y}}\xspace}
\newcommand{\Z}{\ensuremath{\mathbf{Z}}\xspace}
\newcommand{\D}{\ensuremath{\mathcal{D}}\xspace}
\newcommand{\Hyp}{\ensuremath{\mathcal{H}}\xspace}
\newcommand{\Pow}{\ensuremath{\wp}\xspace}

\newcommand{\limply}{\ensuremath{\rightarrow}\xspace}
\newcommand{\lequiv}{\ensuremath{\leftrightarrow}\xspace}
\newcommand{\indicator}{\ensuremath{\mathds{1}}\xspace}

\usepackage[accepted]{icml2022}

\usepackage{amsmath}
\usepackage{amssymb}
\usepackage{mathtools}
\usepackage{amsthm}

\usepackage[capitalize,noabbrev]{cleveref}
\theoremstyle{plain}
\newtheorem{theorem}{Theorem}

\theoremstyle{definition}
\newtheorem{definition}[theorem]{Definition}

\theoremstyle{remark}

\usepackage[textsize=tiny]{todonotes}

\icmltitlerunning{Learning Modulo Theories}

\usepackage{amsmath,amsfonts,bm}

\def\eqref#1{equation~\ref{#1}}

\def\1{\bm{1}}

\DeclareMathAlphabet{\mathsfit}{\encodingdefault}{\sfdefault}{m}{sl}
\SetMathAlphabet{\mathsfit}{bold}{\encodingdefault}{\sfdefault}{bx}{n}

\DeclareMathOperator*{\argmax}{arg\,max}
\DeclareMathOperator*{\argmin}{arg\,min}

\begin{document}
\twocolumn[
\icmltitle{Learning Modulo Theories}



\icmlsetsymbol{equal}{*}

\begin{icmlauthorlist}
\icmlauthor{Matt Fredrikson}{cmu}
\icmlauthor{Kaiji Lu}{cmu}
\icmlauthor{Saranya Vijayakumar}{cmu}
\icmlauthor{Somesh Jha}{uwm}
\icmlauthor{Vijay Ganesh}{uw}
\icmlauthor{Zifan Wang}{cmu}
\end{icmlauthorlist}

\icmlaffiliation{cmu}{Carnegie Mellon University, Pittsburgh, PA, USA}
\icmlaffiliation{uwm}{University of Wisconsin-Madison, Madison, WI, USA}
\icmlaffiliation{uw}{University of Waterloo, Waterloo, ON, Canada}

\icmlcorrespondingauthor{Matt Fredrikson}{mfredrik@cmu.edu}

\icmlkeywords{Machine Learning, ICML}

\vskip 0.3in
]



\printAffiliationsAndNotice{}  


\begin{abstract}
Recent techniques that integrate \emph{solver layers} into Deep Neural Networks (DNNs) have shown promise in bridging a long-standing gap between inductive learning and symbolic reasoning techniques.
In this paper we present a set of techniques for integrating \emph{Satisfiability Modulo Theories} (SMT) solvers into the forward and backward passes of a deep network layer, called \layername.
Using this approach, one can encode rich domain knowledge into the network in the form of mathematical formulas.
In the forward pass, the solver uses symbols produced by prior layers, along with these formulas, to construct inferences; in the backward pass, the solver informs updates to the network, driving it towards representations that are compatible with the solver's theory.
Notably, the solver need not be differentiable.
We implement \layername as a Pytorch module, and our empirical results show that it leads to models that \emph{1)} require fewer training samples than conventional models, \emph{2)} that are robust to certain types of covariate shift, and \emph{3)} that ultimately learn representations that are consistent with symbolic knowledge, and thus naturally interpretable.
\end{abstract}


\section{Introduction}

A recent class of techniques aims at integrating {\it solver layer(s)} within deep neural networks (DNNs)~\cite{satnet19,blackboxsolver20,huang2021scallop,manhaeve2018deepproblog}, both during training and inference.
A class of problems which can benefit from such an integration is one that has both a perceptual and a symbolic sub-problem, such as ``visual'' Sudoku~\cite{satnet19}, or the problem of determining the shortest path from a picture of a map~\cite{blackboxsolver20}.

The most straightforward way to incorporate a solver layer into an ML model is to learn models with representations that are compatible with symbols used by the solver.
For example, if one wanted to leverage symbolic domain knowledge to classify images of birds, or diagnose ailments from CT scans, then one could train a model in a fashion similar to ``concept bottlenecking''~\cite{koh2020concept}.
This requires detailed labels for supervision, which may be prohibitively expensive to obtain and keep consistent with a potentially evolving domain theory.

We present a set of techniques for incorporating a \emph{Satisfiability Modulo Theories} (SMT) solver into a DNN layer so that symbolic knowledge can be leveraged to learn such a compatible representation, \emph{without requiring label supervision}. 
Our approach is general, and can handle a broad range of domain knowledge encoded as SMT constraints, provided that they interface with the surrounding neural network layers over propositional variables.
Unlike the most closely related prior work~\cite{satnet19}, our approach does not approximate the solver's behavior by formulating a differentiable relaxation.
Rather, we extract information from the solver as it works on a set of constraints, that is geared towards checking the correctness of the output of the model that precedes the solver, and use that information to construct updates to the model during training (Section~\ref{sect:layer-details}).

We present two different approaches for this, one based on unsatisfiable cores, and another based on weighted MaxSMT (Section~\ref{sect:layer-details}).
There are several advantages to this approach.
Aside from the mild interface constraints mentioned earlier (i.e., solver and neural layers interface with each other via boolean variables), our approach does not place any restrictions on the theory solver embedded in the layer, such as linearity~\cite{blackboxsolver20} or even decidability---if the solver is capable of efficiently discharging the relevant constraints, then the layer can operate as intended.
Because there is no need to provide a differentiable relaxation for each theory or solver technique that one may want to incorporate, we can leverage the continuous and unabated progress being made in solver technology.

We implement our approach as a PyTorch~\cite{pytorch} layer, using the Z3~\cite{Z3-paper} SMT solver as the solver layer to solve SMT and MaxSMT constraints.
On three applications involving vision and natural language: visual arithmetic, algebraic equation solving, and a so-called natural language ``liar's puzzle,'' we demonstrate that our implementation can be incorporated into DNN architectures to solve problems more effectively than conventional DNNs (Section~\ref{sect:eval}).
In particular, our results show that the data needed to train a DNN with symbolic knowledge may be much simpler than may be necessary otherwise, and that while doing so is more expensive computationally, often times the more efficient (i.e., not involving MaxSMT) algorithms perform well in practice.

Our contributions are as follows:
\begin{enumerate}

\item We present \layername, a framework for incorporating an SMT solver into a DNN, as a layer that leverages symbolic knowledge during training and inference.

\item We prototype our approach in Pytorch\footnote{We plan to release our implementation as an open source library upon publication of this paper}, and show that it can be applied to solve a range of problems that incorporate symbolic knowledge.

\item Our empirical evaluation, over four diverse applications, shows that models using \layername require significantly less training data, can be trained more efficiently, and are more robust than those based on closely-related prior work~\cite{satnet19, huang2021scallop}.

\end{enumerate}

Section~\ref{sect:background} provides background on ERM and the first-order theories used in our framework.
Section~\ref{sect:layer} describes \layername, Section~\ref{sect:eval} gives our empirical evaluation, and Section~\ref{sect:conclusion} concludes the paper.

\section{Related Work}
\label{sect:related}

Combining logical solvers and deep models can be difficult because logic has discrete structure while the most successful way to construct neural networks today requires differentiability~\cite{riegel2020logical}.



\paragraph{Combinatorial Solver Layers.} \citet{vlastelica2019differentiation} integrate a blackbox, non-differentiable combinatorial solver on top of a deep network. To propagate the gradient through the solver on the backward pass, they linearly interpolate the loss w.r.t the solver's input and define the gradient of the solver as the slopes of the line segments. CSL solves a set of problems where the solver's objective must be linear w.r.t its input, e.g. finding the shortest path and travel salesman problem (TSP). Further, the authors assume that the only labels available are the outputs of solvers, e.g. the minimum cost in TSP, and hence their tool has to discover the label for the output of the network itself. These requirements limit the choices one has for the solver layer.

\paragraph{Neural Logic Programming.} 
While SATNet integrates a logic-based solver on top of a network, DeepProbLog takes the opposite approach, extending the capability of a probabilistic logic programming language with neural predicates \citet{manhaeve2018deepproblog}.
In the context of our work, the logic program can be viewed as a ``solver layer'' that explicitly encodes symbolic knowledge.
Scallop~\cite{huang2021scallop} extends DeepProbLog to scale without sacrificing accuracy compared to DeepProbLog. Similarly to DeepProbLog, each possible result of the sum of two digits in MNIST is given a probability, in the form of a weighted Boolean formula. They prune unlikely clauses of the formula, represented by proofs, only keeping the top-$k$ most likely. Likelihood is computed using weighted model counting~\cite{huang2021scallop, chavira2008probabilistic}.
These techniques are well-suited to problems that benefit from probabilistic Datalog, but have inherent limitations: they cannot handle quantifiers, general negation, and the range of supported first-order theories is more restrictive.

\paragraph{SATNet.} 
\citet{wang2019satnet} present SATNet, a network architecture with a differentiable approximate MAXSAT solver layer. Their approximation is based on a coordinate descent approach to solving the semidefinite program (SDP) relaxation of the MAXSAT problem. 
SATNet does not assume that the logical structure of the problem is given, and instead attempts to learn it. 
By placing the MAXSAT solver layer on top of a convolution network to learn representations from images, SATNet directly solve problems like Visual Sudoku, for which neural networks alone are not well suited~\cite{wang2019satnet}.

\paragraph{Differentiable Logic.}
Another recent direction has explored differentiable logics~\cite{dl2_useless, stl_difflogic, fuzzydiff_whocares}.
These approaches provide ways of integrating symbolic knowledge into training, by making logical formulas differentiable, and therefore amenable to optimization when included in a loss function.
This line of work does not explicitly aim to make use of symbolic information during inference.
In contrast, the information that our approach extracts from the solver during training is used to condition the model towards a representation that will allow it to communicate effectively with the solver during inference.
Additionally, we do not require the logical formulas, or the solver, to be differentiable.



\section{Background}
\label{sect:background}



Let \X denote a domain of features, \Y a domain of labels, and \D a distribution over $\X \times \Y$.
Formally, \D is a probability measure on a space given by a $\sigma$-algebra over subsets of $\Pow(\X \times \Y)$.
The goal of a learning algorithm $A$ is to find a function $h: \X \to \Y$ that, for $(x, y) \sim \D$, can be used to predict $y$ when given $x$.
To do this, $A$ is given a set of training examples $S = (x_1, y_1), \ldots, (x_m, y_m)$ sampled i.i.d. from \D, and uses some criterion to select $h$ from a hypothesis class \Hyp of functions.
We refer to $h$ as the \emph{model} learned by $A$ on $S$.
When the learning algorithm $A$ is clear from the context, we will write $h_S$ to denote the model produced from the given sample.
Throughout this paper, we will generally assume that the loss is either the 0-1 loss $\ell^{01}$ or binary cross-entropy $\ell^{\mathrm{bce}}$.



A theory $\T$ consists of a signature $\Sigma$ of constant, predicate, and function symbols, as well as a set of axioms over $\Sigma$. 
Formulas in a theory are composed of elements of $\Sigma$, variables, and logical symbols such as quantifiers and Boolean operations. 
We use the term \emph{decision procedure} to refer to an algorithm that is given an open \T-formula, and returns \emph{true} if it is satisfiable, and \emph{false} otherwise.
Additionally, it may return an assignment to all of the variables that demonstrates satisfiability, or if the formula is not satisfiable, then it may return an \emph{unsatisfiable core}, which is a subset of clauses taken from the formula's representation in conjunctive normal form that remains unsatisfiable.
Loosely, we also refer to such an algorithm as a ``solver'', but this term is more general, and could also refer to an algorithm that identifies the maximal set of clauses, possibly weighted by some user-defined values, that are satisfiable when conjoined.



\section{Constructing \layername}
\label{sect:layer}



In this section, we present \layername, a set of algorithms for computing the forward and backward passes of a layer whose behavior is defined by a set of user-defined SMT constraints. \layername does not have trainable parameters, and its functionality is wholly defined by a set of SMT constraints $\phi$ that are provided by the model designer. \layername can be used in modern deep-learning frameworks as a drop-in replacement for more conventional neural network layers, e.g., dense, convolutional, and LSTM~\cite{10.1162/neco.1997.9.8.1735} are prominent examples of widely-used layers.



Section~\ref{sect:layer_overview} provides a high-level overview of our approach, Section~\ref{sect:layer-details} describes them in detail, and Section~\ref{sect:layer-analysis} begins an analysis of this setting that we hope future work will continue developing.


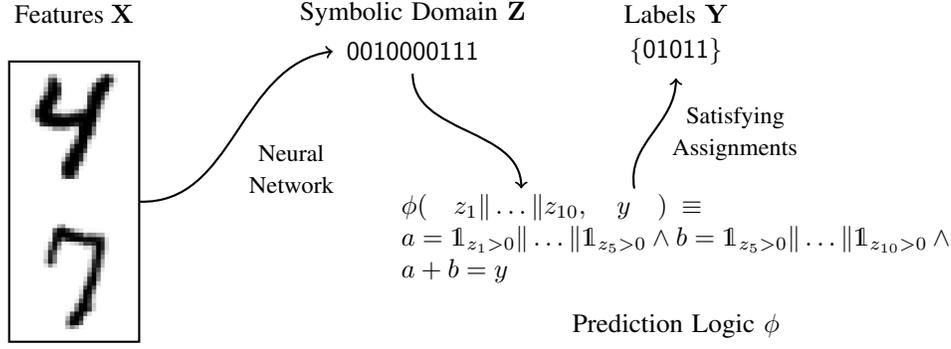
\begin{figure*}[t]
\begin{center}
\begin{tikzpicture}[
label/.style 2 args={
  postaction={
    decorate,
    transform shape,
    decoration={
      pre length=1pt, post length=1pt,
      markings,
      mark=at position #1 with \node #2;
      }
  }
}  
]

\node[inner sep=0pt] (features) at (-0.5,1.5) 
	{Features \X};
\node[inner sep=0pt] (four) at (-0.5,0)
    {\includegraphics[width=.1\textwidth]{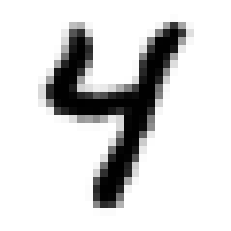}};
\node[inner sep=0pt] (seven) at (-0.5,-2)
    {\includegraphics[width=.1\textwidth]{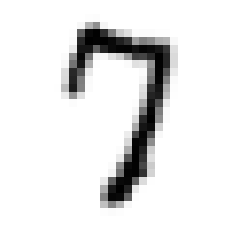}};
\draw[thick] ($(four.north west)$) rectangle ($(seven.south east)$);

\node[inner sep=0pt,text width=3cm,align=center] (zdomain) at (4,1.5) 
	{Symbolic Domain \Z};
\node[inner sep=5pt] (bv_rep) at (4,1) 
{$\mathsf{0010000111}$};

\draw[->,thick,bend right] ($(four.north east)!0.5!(seven.south east)$) 
							to [out=320,in=150] 
							node [pos=0.55,yshift=-4ex,xshift=4ex,align=center,text width=3cm] 
								{\small Neural \\ Network}
							(bv_rep.west);

\node[inner sep=1pt] (logic) at (7.5,-1.5) 
	{
	$\arraycolsep=1pt
	\begin{array}{ll}
		\phi(\quad z_1\|\ldots\|z_{10}, \quad y \quad)\ \equiv & \\
		a = \indicator_{z_1 > 0}\| \ldots \| \indicator_{z_5 > 0} \land 
		b = \indicator_{z_5 > 0} \| \ldots \| \indicator_{z_{10} > 0} & \land \\
		a + b = y &
	\end{array}
	$};
\node[inner sep=5pt,text width=3cm,align=center,below of=logic,yshift=-1ex] (f) {Prediction Logic $\phi$};

\node[inner sep=0pt,text width=3cm,align=center] (f) at (7.5,1.5) 
	{Labels \Y};
\node[inner sep=5pt] (result) at (7.5,1) 
	{$\{\mathsf{01011}\}$};

\draw[->,thick] (bv_rep.south) to [out=270,in=90] ([xshift=-13ex]logic.north);
\draw[->,thick] ([xshift=-3.5ex]logic.north) to [out=100,in=270] 
				node [pos=0.5,xshift=7ex,align=center,text width=3cm] {\small Satisfying \\ Assignments} 
				(result.south);
\end{tikzpicture}
\end{center}
\caption{\label{fig:mnist-example} MNIST Addition example.}
\end{figure*}

\subsection{Overview}
\label{sect:layer_overview}

We envision \layername being used primarily at the top of a DNN taking inputs from a stack of conventional DNN layers that convert raw input features into ground terms for the constraints $\phi(z_0, \ldots, z_{p-1}, y_0, \ldots, y_{q-1})$ embedded in \layername, and producing outputs that are consistent with $\phi$ and the given ground terms.
Figure~\ref{fig:mnist-example} shows an illustrative example, with the previously-studied problem of MNIST addition~\cite{manhaeve2018deepproblog,huang2021scallop}.

During the forward pass the outputs of the previous layer are mapped to designated free variables $z_0, \ldots, z_{p-1}$.
The layer then checks the satisfiability of $\phi$, a formula in an appropriate combination of first-order theories, after substituting these ground terms for the $z_i$, and the output of the layer consists of the solver's model for $y_0, \ldots, y_{q-1}$. 
These outputs are converted from Boolean to floating-point values by mapping \emph{false} to -1 and \emph{true} to 1.
At the moment, the only restriction on $\phi$ that our layer requires is that $z$ and $y$ be vectors of Booleans, so that they can be appropriately mapped to continuous values; any other symbols appearing in $\phi$ can come from arbitrary domains (e.g. strings) supported by the underlying SMT solver.

In the backward pass, the layer receives the gradient of its output with respect to the function whose derivative is being computed, which we will assume is the binary cross-entropy loss $\ell(y, y^\star)$.
Unless stated otherwise, we will assume this loss for the remainder of the section.
This gradient is used, along with the inputs and outputs of the corresponding forward pass, to first compute an amended output $\hat{y}$ which corresponds to an output that would have yielded a smaller loss.
Because the outputs are Boolean, it is always possible to determine the ground truth output $y^\star$ from this information.
Using $\hat{y}$, the layer determines which of components of its inputs are inconsistent with $\phi$ and $\hat{y}$, and provides the corresponding gradients to the previous layer.
Section~\ref{sect:layer-details} details the manner in which these gradients are computed.

\subsection{\layername, forward and backward}
\label{sect:layer-details}

\begin{algorithm}[t]
    \small
    \DontPrintSemicolon
    \SetAlgoLined
    \LinesNumbered
    \setcounter{AlgoLine}{0}
    \KwIn{$z \in \mathbb{R}^p$ layer input \newline 
          $\phi(z_0, \ldots, z_{p-1}, y_0, \ldots, y_{q-1})$ \T-formula }
    \KwOut{$y \in \mathbb{R}^q$}
    
    \Begin{
        $z_b \gets [z[i] > 0~:~i = 0 \ldots p-1]$

        $C \gets \sum_{i \in I} \mathsf{softmax}(|z|)[i]$
        
        $y \gets \argmax_{y_b} \max_{I} C \cdot \indicator(\phi \land \bigwedge_{i \in I} y_i = y_b[i] \land z_i = z_b[i])$

        \Return $y$ \;
    }
    \caption{$\forwardmax^\phi(z)$\\MaxSMT-based forward pass of \layername\label{alg:forward-max}}
\end{algorithm}

\begin{algorithm}[t]
    \small
    \DontPrintSemicolon
    \SetAlgoLined
    \setcounter{AlgoLine}{0}
    \KwIn{$z \in \mathbb{R}^p$ layer input \newline 
          $\phi(z_0, \ldots, z_{p-1}, y_0, \ldots, y_{q-1})$ \T-formula }
    \KwOut{$y \in \mathbb{R}^q$}
    
    \Begin{
        $z_b \gets [z[i] > 0~:~i = 0 \ldots p-1]$
        
        $\hat{\phi} \gets \phi(z_b[0], \ldots, z_b[p-1])$
        
        \eIf{$\hat{\phi}$ is satisfiable}{
            $y_b[0], \ldots, y_b[q-1] \gets \mathsf{solve}(\hat{\phi}, y_0, \ldots, y_{q-1})$ 
            
            $y \gets [y_b[i] > 0 \quad~:~i = 0 \ldots q-1]$ \;
        }{
            $y \gets \mathbf{0}$
        }
        \Return $y$ \;
    }
    \caption{$\forward^\phi(z)$ \\ SMT-based forward pass of \layername\label{alg:forward}}
\end{algorithm}

We now present the details of the forward and backward passes of \layername.
There are two algorithms for each pass, $\forwardmax^\phi$ and $\forward^\phi$ are forward passes, and $\backmax^\phi$, $\backcore^\phi$ are backward passes.
$\forwardmax^\phi$ and $\backmax^\phi$ both make use of MaxSMT solvers, whereas $\forward^\phi$ and $\backcore^\phi$ rely on satisfiability solvers (SMT).
Despite the symmetry in which type of solver each algorithm uses, they are all compatible with each other.
That is, $\forward^\phi$ can be used with either $\backmax^\phi$ or $\backcore^\phi$, and the same for $\forwardmax^\phi$.

\paragraph{Forward pass.}
Algorithms~\ref{alg:forward-max} and \ref{alg:forward} illustrate $\forwardmax^\phi$ and $\forward^\phi$, the methods for computing the forward pass based on weighted MaxSMT and SMT, respectively.
Both of the algorithms are parameterized by a user-provided first-order formula $\phi$, and take a single vector-valued input consisting of unscaled floating-point values (\emph{logits}).
These values are cast to Boolean constants by taking their sign on line 2 of both algorithms, so that they can be equated with the corresponding free variables $z_0, \ldots, z_{p-1}$.

The key difference between $\forwardmax^\phi$ and $\forward^\phi$ is the way in which they handle inputs that are inconsistent with $\phi$ when interpreted as Booleans.
$\forward^\phi$ addresses this by providing an output that is also inconsistent with $\phi$, i.e. a vector of zeroes, effectively signaling that the network below it did not provide consistent inputs.
Alternatively, we can interpret the values provided by the network as Booleans enriched with ``confidence'' values.
Although we expect inputs to \layername to be unscaled floating-point values, Algorithm~\ref{alg:forward-max} scales them to a formal probability distribution via the softmax function (line 3) for use as weights to find the weighted MaxSMT solution of $\phi$.
With this approach, \layername will always provide a valid, although not necessarily correct, output that is consistent wrt $\phi$ with the inputs of which the network below is most ``confident.'' (line 4).

\begin{algorithm}[t]
    \small
    \DontPrintSemicolon
    \SetAlgoLined
    \setcounter{AlgoLine}{0}
    \KwIn{$z \in \mathbb{R}^p$ input of forward pass \newline 
          $y \in \mathbb{R}^q$ output of forward pass \newline 
          $\partial_y \ell(y,y^\star)$ gradient with respect to output \newline 
          $\phi(z_0, \ldots, z_{p-1}, y_0, \ldots, y_{q-1})$ a \T-formula }
    \KwOut{$\partial_z \ell(y,y^\star) \in \mathbb{R}^p$ approximate gradient of $\ell$}
    
    \Begin{
        $G_z \gets \partial_z \ell(z, \mathsf{sign}(z))$
        
        $\hat{y} = \mathsf{sign}(y) - 2\cdot\mathsf{sign}\left(\partial_y \ell(y,y^\star)\right)$ 
        
        \If{$\mathsf{sign}(y) \ne \mathsf{sign}(\hat{y})$}{%
            $z_b \gets [z[i] > 0~:~i = 0 \ldots p-1]$

            $\hat{y}_b \gets [\hat{y}[i] > 0~:~i = 0 \ldots q-1]$
            
            $\phi_y \gets \bigwedge_{0 \le i < q} y_i = \hat{y}_b[i]$

            $C \gets \sum_{i \in I} \mathsf{softmax}(|z|)[i]$
            
            $I \gets \argmax_{I \subseteq [0, p)} \indicator(\phi \land \phi_y \land \bigwedge_{i \in I} z_i = z_b[i]) \cdot C$
            
            \ForEach{$i \in \bar{I}$}{$G_z[i] \gets \partial_{z[i]} \ell(z[i], 1 - \mathsf{sign}(z[i]))$}
        }
        \Return $G_z$ \;
    }
    \caption{$\backmax^\phi(z, y, \partial_y\ell(y,y^\star))$\\MaxSMT-based backward pass of \layername\label{alg:max-backprop}}
\end{algorithm}

\begin{algorithm}[t]
    \small
    \DontPrintSemicolon
    \SetAlgoLined
    \setcounter{AlgoLine}{0}
    \KwIn{$z \in \mathbb{R}^p$ input of forward pass \newline 
          $y \in \mathbb{R}^q$ output of forward pass \newline 
          $\partial_y \ell(y,y^\star)$ gradient with respect to output \newline 
          $\phi(z_0, \ldots, z_{p-1}, y_0, \ldots, y_{q-1})$ a \T-formula }
    \KwOut{$\partial_z \ell(y,y^\star) \in \mathbb{R}^p$ approximate gradient of $\ell$}
    
    \Begin{
        $G_z \gets \partial_z \ell(z, \mathsf{sign}(z))$
        
        $\hat{y} = \mathsf{sign}(y) - 2\cdot\mathsf{sign}\left(\partial_y \ell(y,y^\star)\right)$ 

        \If{$\mathsf{sign}(y) \ne \mathsf{sign}(\hat{y})$}{%
            $z_b \gets [z[i] > 0~:~i = 0 \ldots p-1]$ 

            $\hat{y}_b \gets [\hat{y}[i] > 0~:~i = 0 \ldots q-1]$
            
            $\phi_z, \phi_y \gets \bigwedge_{0 \le i < p} z_i = z_b[i],\,\,\, \bigwedge_{0 \le i < q} y_i = \hat{y}_b[i]$ 
            
            $I \gets \argmin_{I \subseteq [0, p)} \indicator(\lnot\phi \lor \lnot\phi_y \lor \bigvee_{i \in I} z_i \ne z_b[i]) \cdot |I|$ 
            
            \ForEach{$i \in I$}{$G_z[i] \gets \partial_{z[i]} \ell(z[i], 1 - \mathsf{sign}(z[i]))$}
            \ForEach{$i \in \bar{I}$}{$G_z[i] \gets 0$}
        }
        \Return $G_z$ \;
    }
    \caption{$\backcore^\phi(z, y, \partial_y\ell(y,y^\star))$\\ unsat core-based backward pass of \layername\label{alg:core-backprop}}
\end{algorithm}

\paragraph{Backward pass.}

The backward pass is responsible for computing the gradient of the loss with respect to the layer inputs.
It is given the gradient of the loss with respect to the layer outputs, and is assumed to have memoized the inputs that it received previously on the forward pass, as well as the outputs that they produced.
The gradients returned by this pass are then used by the backward pass of previous layers, and ultimately to derive updates to trainable parameters that will yield smaller loss.

The key issue in designing a backward pass for \layername is the geometry of the functions computed by either forward pass. 
For any vector $v \in \{-1,0,1\}^p$ and $x, x'$ with $\mathsf{sign}(x) = \mathsf{sign}(x') = v$, then $\mathsf{B}_\cdot^\phi(x) = \mathsf{B}_\cdot^\phi(x')$, so these functions are piece-wise constant step functions ranging over the corners of the $\mathbb{R}^q$ unit hypercube.
Thus, while they are differentiable almost everywhere, the gradient is not helpful for training because it is always zero.
Prior work on integrating such step functions into deep networks primarily addresses this problem by relaxing the function computed by the forward pass, so that its gradients are no longer constant and hopefully more informative.

In contrast, $\backmax^\phi$ (Algorithm~\ref{alg:max-backprop}) and $\backcore^\phi$ (Algorithm~\ref{alg:core-backprop}) do not attempt to provide gradients for a relaxation of the forward pass.
Instead, they use information provided by the solver in its computation of the forward pass to identify which components of the input may have contributed to higher loss.
The gradient is then computed by constructing a counterfactual variant of the input provided to the forward pass, which differs on the identified components, and returning the gradient of the binary cross-entropy loss of the original input on this counterfactual variant.
The two algorithms differ in the information that they extract from the solver, i.e., either solutions to a MaxSMT instance or an unsatisfiable core.

Both algorithms begin by initializing the gradient to be the loss between the logit inputs, and their hard labels (line 2).
Recall that we assume the loss $\ell$ is binary cross-entropy, so the result will not be zero.
The purpose of this initialization is to emulate the dynamics of training with cross-entropy loss with a conventional layer; when the rounded output matches the target, the loss is not zero, and training will continue to move the parameters in a direction that makes them agree ``more'' with the hard target.
One line 3, they then use the provided gradient from the subsequent layer together with the memoized output from the forward pass to construct $\hat{y}$, a ``corrected'' output that satisfies $\ell(\hat{y}, y^\star) \le \ell(y, y^\star)$.
If the sign of $\hat{y}$ is the same as that of $y$, then both algorithms return the initialized gradient.
Otherwise, they extract information from the solver using the inputs to the forward pass $z$ and $\hat{y}$.

$\backmax^\phi$ constructs a set of clauses $\phi_y$ that constrain the free $y_0, \ldots, y_{q-1}$ to take the values of $\hat{y}_b$, the Boolean conversion of $\hat{y}$.
It then computes the softmax values of the absolute incoming logits $|z|$, and uses them to find the maximally-weighted set of clauses $(\mathsf{softmax}(|z|)[i], z_i = z_b[i])$ that are consistent with $\phi \land \phi_y$.
Intuitively, these are the inputs that the previous layer is most confident in that can be made consistent with the corrected label $\hat{y}$ by changing some of the less confident inputs.
$\backmax^\phi$ then updates the initialized gradient at each index for which the solution to this instance does not match the sign of the original input.

$\backcore^\phi$ also constructs $\phi_y$, but instead identifies a set of constraints $z_i = z_b[i]$ that are inconsistent with $\phi \land \phi_y$.
Note that line 7 specifies a minimal unsatisfiable core, but this is not necessary.
All that is needed is that none of the clauses in the core be superfluous, i.e., deleting any singleton clause from $I$ will cause it to be satisfiable.
If a superfluous clause remains in the core, then the gradient returned for the corresponding input will have the incorrect sign, which may lead to issues with training.
$\backcore^\phi$ then updates the gradient at each input identified in the core using the loss of $z[i]$ with respect to $1 - \mathsf{sign}(z[i])$, which will lead to updates in a direction that would have modified the input such that $i$ was not in the unsat core.
The indices not in the unsat core have their gradients set to zero, as their absence in the core is not evidence that these inputs were correct or incorrect.

\subsection{Analysis}
\label{sect:layer-analysis}
\label{sect:theory-basics}

To understand the settings where \layername will provide optimal results, we introduce a class of \emph{decomposable} learning problems (Definition~\ref{def:decomposable}).

\begin{definition}[Decomposable problem]
\label{def:decomposable}
Let \T be a first-order theory with constants in \Z.
An ERM problem \D, \Hyp is \emph{decomposable} by \T if there exists a function $f: \X \to \Z$, companion hypothesis class $\Hyp_f \subseteq \X \to \Z$, and \T-formula $\phi$ such that:
\begin{enumerate}

	\item For any $h \in \Hyp$, there exists $h_f \in \Hyp_f$ and $h'$ such that $h = h' \circ h_f$.

	\item There exists a random function $g : \Pow(\Y) \to \Y$ such that for any $n > 0$ and $\forall S$ in the support of $\D^n$,
	\[
	\label{eq:decompose}
		\Pr_{(x, y) \sim \D}[(x, y) \in S] = \Pr_{(x, \cdot) \sim \D}[(x, g(\langle x \rangle_{f,\phi})) \in S]
	\]
	where $\langle x\rangle_{f,\phi} = \{y : \phi(f(x), y)~\text{is satisfied}\}$.
\end{enumerate}
In (\ref{eq:decompose}), $f$ is called the \emph{grounding function} and $\phi$ is called the \emph{prediction logic}.
\end{definition}

Intuitively, a learning problem defined in terms of a distribution \D and hypothesis class \Hyp is decomposable if members of \Hyp can be decomposed into functions that are responsible for grounding and prediction, and \D can be expressed in terms of a grounding function and a first-order formula $\phi$.
There are a few important things to note.
First, there is no requirement that the grounding function $f$ be a member of $\Hyp_f$.
While this may be realized at times, we should not assume that the data is actually generated, or otherwise described with perfect fidelity, by a function in the class that one learns over.
In fact, we do not assume that $f$ is efficiently computable, as it may correspond to a natural process, or an aspect of data generation that is not understood well enough to make such computational claims.

Second, for a given $x$, there may be more than one satisfying assignment for $y$ to $\phi(f(x), y)$.
The function $g$ in (2) accounts for this, requiring only that when solutions to $\phi(f(x), y)$ are sampled by $g$, the result is distributed identically to \D.
This paper will focus on the case where satisfying assignments for $y$ are unique, as these are more in line with "classic" ERM classification problems.
We leave exploration of the more general setting to future work.

We note that if the grounding function is known, can be computed efficiently, and $\phi$ is efficiently solvable, then the learning problem effectively has a closed-form solution.
Rather, we assume that only $\phi$ and perhaps $g$ are known, and a sample of \D is given.
The remaining challenge is to identify a grounding hypothesis $h_f \in \Hyp_f$ for which the construction in (2) is an effective solution to the end-to-end learning problem posed by \D, \Hyp.
This stands in contrast to traditional ERM, in which a good solution $h \in \Hyp$ must either solve both grounding and prediction, or find a ``shortcut'' that manages to predict \D as well as the decomposition.

\paragraph{Convergence.}

Regarding the backward passes, Theorem~\ref{thm:maxsat-converge} below demonstrates that when $\phi$ satisfies certain conditions, and the companion hypothesis class $\Hyp_f$ satisfies conditions that are sufficient to guarantee convergence with SGD, then training with $\forward^\phi$ and $\backmax^\phi$ will converge to the optimal solution in the number of iterations.
The proof of this theorem is based on the observation that when the conditions on $\phi$ are met, then training with $\backmax^\phi$ obtains the same solution that would be obtained if the labels of $\phi$ were available for supervised learning.
Thus, the conditions on $\Hyp_f$ are sufficient to ensure the stated convergence, as stated in a well-known result outlined in Chapter 14 of~\cite{shalev-shwartz-2014}.

It is also worth noting that Theorem~\ref{thm:maxsat-converge} does not necessarily hold if $\backcore^\phi$ is used instead of $\backmax^\phi$.
The reason is that there may be many unsatisfiable cores that are locally minimal in cardinality, and gradients are set only for inputs that appear in the computed core.
These gradients will not match those of the loss on a grounding sample, so the training dynamics are likely to be different.
We believe that training with $\backcore^\phi$ may have more in common with block coordinate descent than gradient descent, and save a more detailed exploration of the topic for future work.

\begin{theorem}
\label{thm:maxsat-converge}
Let \D, \Hyp be a \T-decomposable problem with grounding function $f$ and prediction logic $\phi$ where:
\begin{enumerate}
    \item \Z and \Y are Cartesian products of Booleans.
    \item For any $(x, y) \sim \D$ and $y' \ne y$, $\phi(f(x), y')$ is \T-equivalent to false and there is exactly one $z$ such that $\phi(z, y)$ is \T-equivalent to true.
    \item $\Hyp_f$ is a convex set and for all $h_f \in \Hyp_f$, $\|h_f\| \le B$, and the loss $\ell(h_f(\cdot), z)$ is $M$-Lipschitz and convex in $x$ for any fixed $z$.
\end{enumerate}
Then for any $\epsilon > 0$, selecting $h_f$ by minimizing either $L_S(\forward^\phi(h_f(\cdot)))$ with $\tau \ge \nicefrac{M^2B^2}{\epsilon^2}$ iterations of stochastic gradient descent, with gradients provided by $\backmax^\phi$, and learning rate $\eta = \sqrt{\nicefrac{B^2}{M^2\tau}}$ yields a grounding hypothesis $\hat{h_f} \in \Hyp_f$ that satisfies:
$
\mathbb{E}[L_\D(\hat{h_f})] \le \min_{h_f \in \Hyp_f} L_\D(h_f) + \epsilon
$.
The randomness in this expectation is taken over the choices of the SGD algorithm.
\end{theorem}

\section{Experimental Evaluation}
\label{sect:eval}

In this section we present an empirical evaluation of \layername on four learning problems that can be decomposed into perceptual and symbolic subtasks.
Our results demonstrate the following primary findings.
\emph{1)} \layername is effective: on every benchmark, it provides superior results over ``conventional'' learning that takes place without encoded symbolic knowledge.
\emph{2)} \layername has distinct advantages over prior approaches.
Compared with SATNet~\cite{satnet19}, it requires \emph{significantly} less training data to converge, and in all cases yields a more accurate model; compared with Scallop~\cite{huang2021scallop}, it is less computationally expensive, requires less training data, and it is more expressive in terms of the knowledge that it can encode.
\emph{3)} Models trained with \layername may be more robust to certain types of covariate shift that occur relative to the symbolic component of the problem; when \layername succeeds at learning a compatible representation, then it will continue to produce correct inferences provided the perceptual component remains stationary.

\subsection{Datasets}
\label{sect:eval-problems}

Additional details on the datasets and corresponding architectures used in our evaluation can be found in Appendix~\ref{sect:data-details}, and specific hyperparameters used when training on each dataset are in Appendix~\ref{sect:hypers}.

\paragraph{MNIST Addition.} The MNIST addition problem is illustrated in Figure~\ref{fig:mnist-example}, and is similar to the benchmark described by \citet{huang2021scallop}.
For training, we use ``MNIST +$p$\%" to denote a training set of size 60,000 that contains $p\%$ of the possible pairs of digits. So $p=100$ indicates all possible pairs of digits are used, and for $p=10$, we only use pairs of the same digit. We use $p=10, 25, 50, 75$ and $100$ in our experiments. 
In all cases, we use the same test set consisting of instances from all possible pairs of digits.

\paragraph{Visual Algebra.} The task is to solve for the variable $x$ in a graphical depiction of the equation $ax + b = c$, where $a, b$ and $c$ are randomly-chosen numbers, and each symbol is depicted visually using EMNIST~\cite{cohen2017emnist} and HASY graphics~\cite{thoma_martin_2017_259444}.
Similar to MNIST addition, the training sample selects $a$ and $b$ uniformly from pairs of the same digit, and $x$ uniformly from the odd numbers between 0 and 9.
The test sample was generated by sampling $a, b$ uniformly from all pairs of digits, and $x$ from all numbers 0 to 9.

\paragraph{Liar's Puzzle.} The liar’s puzzle is comprised of three sentences spoken by three distinct agents: Alice, Bob, and Charlie.
One of the agents is “guilty” of an unspecified offense, and in each sentence, the corresponding agent either states that one
of the other parties is either guilty or innocent. For example, \textit{``Alice says that Bob is innocent."}
It is assumed that two of the agents are honest, and the guilty party is not. The solution to the problem is an identification of the guilty party. 
A formal characterization of the underlying logic is given in Appendix~\ref{sect:data-details}.
We note that the logic has non-stratified occurrences of negation, so it cannot be encoded with Scallop.
We select a training sample that does not fully specify the logic, so conventional training should be insufficient to identify a good model.

\paragraph{Visual Sudoku.} This task is to complete a $9\times 9$ Sudoku board where each entry is an MNIST digit. 
We use the dataset from the SATNet evaluation~\cite{satnet19}, and examine three configurations obtained by sampling 10\%, 50\%, and 100\% of the original training set.
Although there are examples of Sudoku solvers implemented as logic programs, we were not able to implement one in Scallop without violating stratified negation.
When calculating accuracy, we check that the \emph{entire} Sudoku board is correct.

\subsection{Setup}
\label{sect:eval-setup}

We implemented a prototype of our approach using Pytorch~\cite{pytorch} and Z3~\cite{Z3-paper}, which will be made available in open-source when this paper is published. 
When training models with \layername, we use SGD with Nesterov momentum at rate $0.9$ and gradient clipping rate 0.1.
Before training a model with \layername (or a comparison technique, unless stated otherwise), we first pre-train the neural network by replacing \layername with a dense network containing one hidden layer of 512 neurons.
This can potentially limit the number of training updates needed at lower layers, but will not result in a model with a representation that is compatible with symbolic knowledge, so further training is needed.
The models labeled ``conventional'' in our evaluation have the same architecture as the one used for pre-training.
Results were averaged over five runs of training.

Our evaluation was performed on a machine with an Intel i9 1050K CPU, 64GB memory, and a GeForce RTX 3080 accelerator running Ubuntu 20.04.4, with CUDA 11.1.0 and cuDNN 8.0.4.
We developed and tested our prototype with Pytorch version 1.7.0a0+7036e91 and Z3 4.8.14, and the results in our evaluation use these versions as well.

\label{sect:eval-results}

\begin{table*}
\small
\centering
\begin{tabular}{l|cccccccc}
\toprule
\toprule
 & \multicolumn{2}{c|}{\emph{Conventional}} & \multicolumn{2}{|c}{\emph{w/} \layername} & \multicolumn{2}{|c}{\emph{w/} SATNet} & \multicolumn{2}{|c}{\emph{w/} Scallop}\\
\emph{configuration} &  \emph{test} & \multicolumn{1}{c|}{\emph{epoch}} & \emph{test } & \multicolumn{1}{c|}{\emph{epoch}} & \emph{test } &\multicolumn{1}{c|}{\emph{epoch}} & \emph{test } & \emph{epoch } \\
 & \emph{acc. (\%)} & \multicolumn{1}{c|}{\emph{ time (sec.)}}  & \emph{acc.(\%)} & \multicolumn{1}{c|}{\emph{ time (sec.)}}  & \emph{acc.(\%)} & \multicolumn{1}{c|}{\emph{ time (sec.)}}  & \emph{ acc. (\%)} & \emph{time (sec.)} \\ \midrule
MNIST+ 10\% & 10.0 & 7.1 & \textbf{98.1} & 75.4 & 10.0 & 31.0 & 33.7 & 96.3 \\
MNIST+ 25\% & 32.5 & 7.1 & \textbf{98.3} & 74.8 & 34.2 & 30.9 & 65.8 & 96.4 \\
MNIST+ 50\% & 51.5 & 7.0 &  \textbf{98.6} & 75.8 & 54.8 & 32.8 & \textbf{98.4} & 96.5 \\
MNIST+ 75\% & 76.1 & 7.0 & \textbf{98.5} & 75.0 & 78.4 & 31.9 & 93.5 & 96.4 \\
MNIST+ 100\% & 98.3 & 7.1 &  \textbf{98.5} & 75.8 & 96.7 & 33.5 & \textbf{98.6} & 96.6 \\
Vis. Alg. \#1 & 24.1 & 13.2 & \textbf{98.2} & 168.2 & 19.6 & 80.1 & 18.7 & 602.8 \\
Vis. Alg. \#2 & \textbf{25.4} & 11.2 & \textbf{25.4} & 127.2 & 18.6 & 52.5 & 21.3 & 636.1 \\
Liar's Puzzle & 54.2 & 3.1 & \textbf{86.1} & 28.7 & 84.6 & 3.0 & --- & --- \\
Vis. Sudoku 10\% & 0.0 & 6.3 & \textbf{66.0} & 135.7 & 0.0 & 9.9 & --- & ---\\
Vis. Sudoku 50\% & 0.0 & 28.3 & \textbf{73.1} & 608.1 & 0.0 & 45.4 & --- & --- \\
Vis. Sudoku 100\% & 0.0 & 26.7 & \textbf{79.1} & 1199.0 & 63.2 & 86.5 & --- & --- \\
\bottomrule
\end{tabular}

\caption{\label{tab:results} Results after training and inference with \layername versus a conventional architecture. We use the publicly-available implementations of SATNet~\cite{satnet19} and Scallop~\cite{huang2021scallop}, with hyperparameters matching those in their code. All \layername test accuracies were measured with the MaxSMT forward pass. Epoch times are averaged over all epochs on which the model was trained. Cells marked --- denote that the problem is not compatible with the approach.}

\end{table*}

\subsection{Results}

\paragraph{Overall performance.}
In terms of accuracy, Table~\ref{tab:results} shows that \layername outperforms both the conventional network and prior work in terms of accuracy, training time, or both, on all configurations.
While training with \layername (or any of the above approaches) is more expensive than conventional, \layername is consistently faster than Scallop (nearly 4$\times$ in the case of visual algebra).
The per-epoch time to train the SATNet models is less expensive than \layername, but this is not always conclusive.
In the case of visual sudoku, the 10\% \layername model achieved superior error rates in 15 epochs, compared with 100 epochs for the 100\% SATNet model; this means that the \layername model took less than one-tenth the amount of time to train.

It is also worth noting that although Theorem~\ref{thm:maxsat-converge} suggests that Algorithm~\ref{alg:max-backprop} might have learning advantages over Algorithm~\ref{alg:core-backprop}, we found this not to be the case on these datasets. All of the results in Table~\ref{tab:results} were trained with Algorithm~\ref{alg:core-backprop}, and test inference was done using Algorithm~\ref{alg:forward-max}.

\paragraph{Training sample size.}
Because \layername encodes explicit knowledge that is essential to correct inference on these datasets, our approach is able to perform well in data-impoverished settings where the training sample is insufficient to fully specify the symbolic component of the learning task.
This is readily apparent across the results in Table~\ref{tab:results}: in the MNIST addition and first visual algebra configuration, \layername yields a model that performs nearly perfectly despite not being given a sufficient sample in most cases.
Because SATNet must learn the symbolic component, it is at a disadvantage, and in these settings performs similarly to a conventional model.
In theory, Scallop should be able to perform as well as \layername, as it also encodes explicit knowledge.
However, it is unable to learn a useful model for either visual algebra configuration, and does not learn the correct representation for MNIST addition until it is exposed to half of the possible digit pairs during training.

\layername does particularly well on the visual Sudoku dataset introduced by \citeauthor{satnet19}.
When trained on just 10\% of the original sample, it learns a function that \emph{exceeds} the performance of the SATNet model by a healthy margin, which continues to grow as it is exposed to more training data.
On the other hand, we found that SATNet failed to converge with less than the full original training sample.

\paragraph{Robustness \& interpretability.}
The reason that \layername is able to perform well, and often near the optimum, in configurations that other approaches perform poorly on, is that it learns a representation that is consistent with the symbolic knowledge encoded in the \layername.
For example, the constraints that we use for MNIST addition, visual algebra, and visual sudoku all encode digits as bitvectors.
In order to make a correct inference, the neural network must learn to encode MNIST digits in their correct bitvector representation. 
If learning succeeds at this, then there are two positive outcomes that follow.
First, the model's representation will be inherently interpretable, because it will coincide with the provided symbolic domain knowledge, which is also (presumably) interpretable.
Second, the resulting model is naturally robust to covariate shift that does not affect the distribution of perceptual data that the network translates into theory symbols, but that does affect the statistics of their composition.

This type of shift is on display in the MNIST 10\% and visual algebra experiments, where at training time, the model only sees pairs of same-numbered digits, and at test time it is exposed to a substantially different distribution of digit pairs or formulas.
We verified this by examining the representations learned by \layername and Scallop on MNIST Addition 10\%; it is unreasonable to expect SATNet to learn an interpretable representation, as it is not provided with an interpretable theory during training.
As expected, \layername produces the correct representation at the rate of accuracy of a typical MNIST model ($\approx 99\%$), whereas Scallop's digit representation was correct roughly 50\% of the time.
However, architecture plays a role in this robustness, as shown in the \layername  results for the second visual algebra configuration.
Because the network is shown the full instance, and not the individual digits, it learns the training bias. 
Despite having access to the symbolic formulas in \layername, it cannot disentangle the perceptual symbols from their covariance.
Understanding this issue is an important direction for future work.



\section{Conclusion}
\label{sect:conclusion}

Our approach for integrating logical theories into deep learning, \layername, provides a pragmatic solution to the problem of incorporating symbolic knowledge into learning for training and inference, which we demonstrate on several problems involving both perceptual tasks---vision and natural language---and logical reasoning.
Notably, we show that models which incorporate symbolic knowledge during training and inference can outperform conventional models as well as prior work in this area, especially in settings where training data is limited.
Continued progress on automated reasoning techniques has played a pivotal role in the development of several fields over the past decades, and our hope is that the contributions in this paper will aid in progress towards realizing their potential in challenges that surpass the capabilities of existing learning techniques.

\bibliography{ref}
\bibliographystyle{icml2022}

\newpage
\appendix
\onecolumn
\section{Appendix}
\label{sect:appendix}

\subsection{Proofs}
\label{sect:proofs}

\setcounter{theorem}{1}
\begin{theorem}
\label{thm:maxsat-converge}
Let \D, \Hyp be a \T-decomposable problem with grounding function $f$ and prediction logic $\phi$ where:
\begin{enumerate}
    \item \Z and \Y are Cartesian products of Booleans.
    \item For any $(x, y) \sim \D$ and $y' \ne y$, $\phi(f(x), y')$ is \T-equivalent to false and there is exactly one $z$ such that $\phi(z, y)$ is \T-equivalent to true.
    \item $\Hyp_f$ is a convex set and for all $h_f \in \Hyp_f$, $\|h_f\| \le B$, and the loss $\ell(h_f(\cdot), z)$ is $M$-Lipschitz and convex in $x$ for any fixed $z$.
\end{enumerate}
Then for any $\epsilon > 0$, selecting $h_f$ by minimizing either $L_S(\forward^\phi(h_f(\cdot)))$ with $\tau \ge \nicefrac{M^2B^2}{\epsilon^2}$ iterations of stochastic gradient descent, with gradients provided by $\backmax^\phi$, and learning rate $\eta = \sqrt{\nicefrac{B^2}{M^2\tau}}$ yields a grounding hypothesis $\hat{h_f} \in \Hyp_f$ that satisfies:
$
\mathbb{E}[L_\D(\hat{h_f})] \le \min_{h_f \in \Hyp_f} L_\D(h_f) + \epsilon
$.
The randomness in this expectation is taken over the choices of the SGD algorithm.
\end{theorem}
\begin{proof}
To prove this result, we introduce the notion of a \emph{grounding sample}.
\setcounter{theorem}{3}
\begin{definition}[Grounding sample]
\label{def:ground-sample}
Let \D, \Hyp be a \T-decomposable problem with grounding function $f$.
The grounding sample $S_f$ for $S \sim \D$ is given by
$[(x_i, f(x_i)) : (x_i, y_i) \in S]$, i.e., tuples that consist of the first element of each instance in $S$ and its image under $f$.
\end{definition}

Now observe that the conditions stated in assumption (3) are sufficient to yield the result if instead of optimizing $L_S(\forward^\phi(h_f(\cdot)))$, we were given the grounding sample $S_f$ and minimized $L_{S_f}(h_f)$ (see~\cite{shalev-shwartz-2014}, Theorem 14.8).
The result follows as stated then because of assumptions (1) and (2), which imply that the update vectors provided by $\backmax^\phi$ are the gradients of $L_{S_f}(h_f)$.

To understand why, observe that the sign of $\hat{y}$ computed on line 3 of both algorithms must be equal to that of $y^\star$.
This follows from two facts:
\begin{enumerate} 
    \item At any coordinate $i$ where $y[i] \ne y^\star[i]$, $\mathsf{sign}(\partial_y\ell(y, y^\star))[i] = \mathsf{sign}(y)[i]$.
    \item At any coordinate $i$ where $y[i] = y^\star[i]$, $\mathsf{sign}(\partial_y\ell(y, y^\star))[i] = -1 \cdot \mathsf{sign}(y)[i]$.
\end{enumerate}
Now there are two cases to consider.
\paragraph{Case 1: $\mathsf{sign}(y) = \mathsf{sign}(\hat{y})$.}
In this case the algorithm returns $\partial_z \ell(z, \mathsf{sign}(z))$. 
Because solutions to $\phi(\cdot, y)$ are unique, $\mathsf{sign}(z)$ is the correct grounding, i.e., $z = f(x)$ for the original features $x$.

\paragraph{Case 2: $\mathsf{sign}(y) \ne \mathsf{sign}(\hat{y})$.}
Because of assumption (2), the set of indices computed on line 8 will contain all of the coordinates at which $z$ matches the correct value $z^\star = f(x)$.
Note that at these coordinates, the vector returned by the algorithm matches the gradient of $\ell(z, z^\star)$, which is $\partial_{z[i]} \ell(z[i], \mathsf{sign}(z[i]))$.
In the remaining coordinates, the vector will contain $\partial_{z[i]} \ell(z[i], 1 - \mathsf{sign}(z[i]))$, which also matches the gradient of $\ell(z, z^\star)$.
The result follows.
\end{proof}

\subsection{Dataset details}
\label{sect:data-details}

Two of the three problems that we examine are based on the MNIST handwritten digit dataset~\cite{mnist}, which consists of 60,000 28x28 gray-scale images of handwritten numerals for training and 10,000 instances for testing.
The digits on the left of Figure~\ref{fig:mnist-example} are examples of instances from this data.
To generate data for the visual algebra problem, we additionally drew from EMNIST~\cite{emnist}, which extends MNIST with handwritten letters, and HASY~\cite{hasy}, which contains handwritten symbols with similar characteristics to MNIST.
For the liar's puzzle, inspired by examples~\cite{yurichev2020sat} which formulate similar examples as SMT constraints, we constructed examples using a set of common phrases that we devised ourselves, and did not otherwise draw from publicly-available data.

Below, we describe the way in which we used these data sources to construct training and test samples, and the neural network architectures that we used with \layername to solve them.

\paragraph{MNIST Addition.}

The MNIST addition problem is described in Example~\ref{fig:mnist-example}.
In each instance, two MNIST digits are presented as features, and the task of the model is to provide their sum represented as a bitvector.
The architecture that we use consists of four convolutional layers with kernels of size 3, depths in the order 64, 64, 128, 128, and a stride of width 2 on the first layer, and two dense layers of width 256 and 4.
This network is applied to each digit, and the results are concatenated to obtain a vector of size 8 that is passed to an instance of \layername with SMT constraints from Example~\ref{fig:mnist-example}, which ultimately produces a vector of width 5 that represents the bitvector sum of the digits.

We generated five training samples starting with one containing only pairs of the same digit, i.e. $(0, 0), (1, 1), \ldots$.
We then added progressively more from the full set of possible pairs, using 25\%, 50\%, and 100\%, and trained on batches of 128 across all datasets.
Note that although we change the number of digit pairs that appear between samples, we always map these pairs to random MNIST images to obtain 60,000 training instances.
This is to ensure that the training sample contains a sufficient sample of MNIST images to be able to perform well on the test data.
In all cases, we use the same test set consisting of instances from all possible pairs of digits.
The purpose of this is to demonstrate that the conventional network will not generalize until it has seen the full distribution, whereas the model with \layername should be able to generalize after seeing many fewer examples.

\paragraph{Visual Algebra.}

The visual algebra problem is described in Example~\ref{fig:visual-alg} and Example~\ref{fig:visual-alg2}.
Recall that features depict handwritten depictions of linear equations of the form $ax + b = c$.
Values for $a, x$ and $b$ are randomly drawn from the range 1-9 to ensure that solutions are unique.
Then the corresponding value of $c$ is decomposed into $c = 10\cdot c_1 + c_2$, and MNIST digits are selected at random to represent $a, b, c_1, c_2$.
A random EMNIST alphabet character is drawn for the variable, and random multiplication, addition, and equality symbols are drawn from HASY.
A minor note is that HASY does not contain the standard equality symbol ``$=$'', so we instead use ``$\doteq$''.
These images are then concatenated horizontally in the appropriate order.

We evaluate two architectures for this problem.
The first uses the same neural network that was used for MNIST addition, and an instance of \layername with the constraints given in Example~\ref{fig:visual-alg}.
It assumes that the four numeric digits in the problem have already been extracted, e.g. by a separate vision routine that can recognize digits from letters and arithmetic symbols, and are provided directly to the model.
The four inputs are given separately to the neural network, which produces four 4-bit bitvectors that are concatenated and passed to \layername, which produces a 4-bit bitvector result.
We refer to this as configuration \#1 in our results.

The second uses an architecture which takes the entire image containing the problem as a whole, and produces a 16-bit bitvector that is passed directly to \layername.
This architecture uses a similar stack of convolutional layers, but has a larger initial dense layer containing 26,112 neurons, as it is given a larger image.
The difference in the convolutional stack is at the second layer, which also has a stride of width 2, to reduce the size of the feature map and mitigate the need for an even larger dense connection.
The difference between these architectures relates to one of the challenges of this problem.
Much of the information contained in the features is irrelevant to the solution, e.g., it is irrelevant which letter is chosen for the variable, or what the arithmetic operators look like, so this architecture must also learn to disregard these parts of the instance.
We refer to this as configuration \#2 in our results and in Example~\ref{fig:visual-alg2}.

We generated a training sample by selecting $a$ and $b$ uniformly from pairs of the same digit, i.e. $(0, 0), (1, 1), \ldots$, and sampling $x$ uniformly from the odd numbers between 0 and 9.
The test sample was generated by sampling $a, b$ uniformly from all pairs of digits, and $x$ from all numbers 0 to 9.
We then map these values to random MNIST, EMNIST, and HASY images to obtain 60,000 samples.
The intention is to study a problem wherein the model is not shown all possible problems (modulo representation as digits), or all of the solutions.
This is more challenging than MNIST addition for two reasons: for a given solution, there are many more compatible ground terms, and the model does not see examples of some of the solutions it must provide for the test set.
Thus, in order for \layername to succeed, it must use the provided symbolic knowledge to approximate the correct grounding function, despite these deficiencies in the data.

\paragraph{Liar's Puzzle.}

The liar's puzzle is comprised of three sentences spoken by three distinct agents: Alice, Bob, and Charlie.
One of the agents is ``guilty'' of an unspecified offense, and in each sentence, the corresponding agent either states that one of the other parties is either guilty or innocent.
It is assumed that two of the agents are honest, and the guilty party is not.
The solution to the problem is an identification of the guilty party. An example is described in Example~\ref{fig:liarspuzzle}

We synthesized a dataset for the liar's puzzle based on a limited set of utterances about who speaks in each sentence, which agent is the subject, and whether the subject is guilty or innocent.
There are five ways of denoting the speaker: \emph{``* says''}, \emph{``* says that''}, \emph{``* said''}, \emph{``* said that''}, and a colon \emph{``* :''} separating the speaker's name from the rest of the sentence.
There are five ways of uttering either innocence or guilt: \emph{``* did it/did not do it''}, \emph{``* is guilty/innocent''}, \emph{``* is definitely guilty/innocent''}, and \emph{``* definitely did it/did not do it''}, \emph{``* is the criminal/is a good person''}.
We generated all of the combinations of subject, speaker, and proclaimed innocence or guilt, and took the product with all possible combinations of these utterances.
The result is a dataset of 375,000 instances, each containing three natural language sentences.

The prediction logic for this problem assumes a set of ground predicates $\mathsf{speaker}(\mathit{agent}, \mathit{sentence})$, $\mathsf{subject}(\mathit{agent}, \mathit{sentence})$, $\mathsf{accuses}(\mathit{sentence})$, and $\mathsf{guilty}(\emph{agent})$.
For example, if the first sentence was \emph{``Alice says that Bob is innocent''}, the ground predicates would be $\mathsf{speaker}(\mathsf{alice}, 1)$, $\mathsf{subject}(\mathsf{bob}, 1)$, and $\lnot\mathsf{guilty}(bob)$.
Then the prediction logic is shown in Equation~\ref{eq:liars-logic}.
\begin{equation}
\label{eq:liars-logic}
\begin{array}{ll}
&
|\{a : \mathsf{guilty}(a) \}| = 1
\\
\land &
\forall a. |\{s : \mathsf{speaker}(a, s) \}| = 1
\\
\land &
\forall s. |\{a : \mathsf{subject}(a, s) \}| = 1
\\ 
\land & 
\forall s, a_1, a_2. \mathsf{speaker}(a_1, s) \land \mathsf{subject}(a_2, s) \limply \\ &\ \ \ \ \textbf{}
	(\lnot\mathsf{guilty}(a_1) \lequiv \mathsf{accuses}(s) \lequiv \mathsf{guilty}(a_2))

\end{array}
\end{equation}
In our implementation, agents and sentence identifiers are encoded unary as 3-bit bitvectors. 
The quantifiers are removed by substituting for each ground term or sentence identifier, and the cardinality constraints are expanded into propositional formulas.
The architecture that we adopt is a two-layer bidirectional long short-term model (LSTM) with 512 dimensions at each layer, and a 300-dimension trainable embedding layer initialized from GloVe-6B~\cite{pennington2014glove}.
The hidden units of the last LSTM layer were connected to 2-layer dense network containing 128 followed by 7 neurons.
This network is applied to each sentence in the input, and the concatenated results are passed to the \layername, which solves the formula in Equation~\ref{eq:liars-logic} to produce a unary encoding of the guilty party.

To evaluate solver layers on this problem, we selected training and test samples by first subsampling half of the full 375,000 available instances.
We then selected half of the $\mathsf{speaker}$, $\mathsf{subject}$, $\mathsf{accuses}$ predicate configurations for all three sentences appearing in this subsample to appear in the training sample, and the other half to appear in the test sample.
To further limit the amount of information in the training sample, we randomly selected one ordering for each predicate configuration to remain for training.
There were 9,400 resulting training instances, and 28,062 test instances.
Restricting the training set as described ensures that the model is trained on a limited subset of possible sentence configurations, and one that is logically disjoint from those that appear in the test sample.
Because there is not enough information in the training sample to learn Equation~\ref{eq:liars-logic}, we expect only the model with \layername to succeed, but to do so it must approximate the grounding function well from a limited sample.

\usetikzlibrary{calc} 
\begin{figure*}[t]
\begin{center}
\begin{tikzpicture}[
label/.style 2 args={
  postaction={
    decorate,
    transform shape,
    decoration={
      pre length=1pt, post length=1pt,
      markings,
      mark=at position #1 with \node #2;
      }
  }
}  
]

\node[inner sep=0pt] (features) at (-0.5,1.5) 
	{Features};
\node[inner sep=0pt] (firstfour) at (-0.5,0)
    {\includegraphics[width=.02\textwidth]{figures/mnist4.png}};
\node[inner sep=0pt] (ex) at (-0.5,-.5)
    {\includegraphics[width=.02\textwidth]{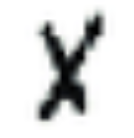}};
\node[inner sep=0pt] (plus) at (-0.5,-1)
    {\includegraphics[width=.02\textwidth]{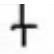}};    
\node[inner sep=0pt] (four) at (-0.5,-1.5)
    {\includegraphics[width=.02\textwidth]{figures/mnist4.png}};
\node[inner sep=0pt] (equals) at (-0.5,-2)
    {\includegraphics[width=.02\textwidth]{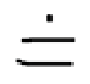}};
\node[inner sep=0pt] (two) at (-0.5,-2.5)
    {\includegraphics[width=.02\textwidth]{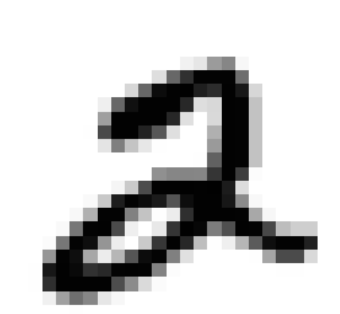}};
\node[inner sep=0pt] (lastfour) at (-0.5,-3)
    {\includegraphics[width=.02\textwidth]{figures/mnist4.png}};

\draw[thick] ($(firstfour.north west)$) rectangle ($(lastfour.south east)$);
\draw[thick] ($(firstfour.north west)$) rectangle ($(firstfour.south east)$);
\draw[thick] ($(firstfour.north west)$) rectangle ($(ex.south east)$);
\draw[thick] ($(firstfour.north west)$) rectangle ($(plus.south east)$);
\draw[thick] ($(firstfour.north west)$) rectangle ($(four.south east)$);
\draw[thick] ($(firstfour.north west)$) rectangle ($(equals.south east)$);
\draw[thick] ($(firstfour.north west)$) rectangle ($(two.south east)$);

\node[inner sep=0pt,text width=3cm,align=center] (zdomain) at (4,1.5) 
	{Symbolic Domain \Z};
\node[inner sep=5pt] (bv_rep) at (4,1) 
{$\mathsf{0010000111}$}; 

\draw[->,thick,bend right] ($(four.north east)!0.5!(seven.south east)$) 
							to [out=320,in=150] 
							node [pos=0.55,yshift=-4ex,xshift=4ex,align=center,text width=3cm] 
								{\small Grounding \\ Function $f$}
							(bv_rep.west);

\node[inner sep=1pt] (logic) at (7.5,-1.5) 
	{
	$\arraycolsep=1pt
	\begin{array}{ll}
		\phi(\quad z_1\|\ldots\|z_{10}, \quad y \quad)\ \equiv & \\
		a = \indicator_{z_1 > 0}\| \ldots \| \indicator_{z_5 > 0} \land 
		b = \indicator_{z_5 > 0} \| \ldots \| \indicator_{z_{10} > 0} & \land \\
		c = \indicator_{z_5 > 0} \| \ldots \| \indicator_{z_{10} > 0} & \land \\
		ax + b = c &
	\end{array}
	$};
\node[inner sep=5pt,text width=3cm,align=center,below of=logic,yshift=-1ex] (f) {Prediction Logic $\phi$};

\node[inner sep=0pt,text width=3cm,align=center] (f) at (7.5,1.5) 
	{Labels \Y};
\node[inner sep=5pt] (result) at (7.5,1) 
	{$\{\mathsf{00101}\}$};

\draw[->,thick] (bv_rep.south) to [out=270,in=90] ([xshift=-13ex]logic.north);
\draw[->,thick] ([xshift=-3.5ex]logic.north) to [out=100,in=270] 
				node [pos=0.5,xshift=7ex,align=center,text width=3cm] {\small Satisfying \\ Assignments} 
				(result.south);
\end{tikzpicture}
\end{center}
\caption{\label{fig:visual-alg} Visual Algebra configuration 1 example.}
\end{figure*}

\begin{figure*}[t]
\begin{center}
\begin{tikzpicture}[
label/.style 2 args={
  postaction={
    decorate,
    transform shape,
    decoration={
      pre length=1pt, post length=1pt,
      markings,
      mark=at position #1 with \node #2;
      }
  }
}  
]

\node[inner sep=0pt] (features) at (-0.5,1.5) 
	{Features};
\node[inner sep=0pt] (formula) at (-0.5,0)
    {\includegraphics[width=.2\textwidth]{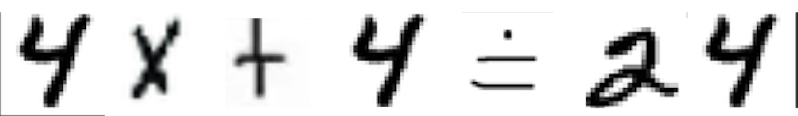}};

\draw[thick] ($(formula.north west)$) rectangle ($(formula.south east)$);

\node[inner sep=0pt,text width=3cm,align=center] (zdomain) at (4,1.5) 
	{Symbolic Domain \Z};
\node[inner sep=5pt] (bv_rep) at (4,1) 
{$\mathsf{0100 0100 0010 0100}$};

\draw[->,thick,bend right] ($(formula.north east)!0.5!(formula.south east)$) 
							to [out=320,in=150] 
							node [pos=0.55,yshift=-4ex,xshift=4ex,align=center,text width=3cm] 
								{\small Grounding \\ Function $f$}
							(bv_rep.west);

\node[inner sep=1pt] (logic) at (7.5,-1.5) 
	{
	$\arraycolsep=1pt
	\begin{array}{ll}
		\phi(\quad z_1\|\ldots\|z_{10}, \quad y \quad)\ \equiv & \\
		a = \indicator_{z_1 > 0}\| \ldots \| \indicator_{z_5 > 0} \land 
		b = \indicator_{z_5 > 0} \| \ldots \| \indicator_{z_{10} > 0} & \land \\
		ax + b = c &
	\end{array}
	$};
\node[inner sep=5pt,text width=3cm,align=center,below of=logic,yshift=-1ex] (f) {Prediction Logic $\phi$};

\node[inner sep=0pt,text width=3cm,align=center] (f) at (7.5,1.5) 
	{Labels \Y};
\node[inner sep=5pt] (result) at (7.5,1) 
	{$\{\mathsf{00101}\}$};

\draw[->,thick] (bv_rep.south) to [out=270,in=90] ([xshift=-13ex]logic.north);
\draw[->,thick] ([xshift=-3.5ex]logic.north) to [out=100,in=270] 
				node [pos=0.5,xshift=7ex,align=center,text width=3cm] {\small Satisfying \\ Assignments} 
				(result.south);
\end{tikzpicture}
\end{center}
\caption{\label{fig:visual-alg2} Visual Algebra configuration 2 example.}
\end{figure*}

\begin{figure*}[t]
\begin{center}
\begin{tikzpicture}[
label/.style 2 args={
  postaction={
    decorate,
    transform shape,
    decoration={
      pre length=1pt, post length=1pt,
      markings,
      mark=at position #1 with \node #2;
      }
  }
}  
]

\node[inner sep=0pt] (features) at (-0.5,1.5) 
	{Input};
\node[draw,align=left] (lies) at (-0.5,0)
    {\textbf{Alice} said that Bob \\ did not do it.\\ \textbf{Bob}: Alice is \\definitely innocent \\\textbf{Charlie}: Alice did it };

\draw[thick] ($(lies.north west)$) rectangle ($(lies.south east)$);

\node[inner sep=0pt,text width=3cm,align=center] (zdomain) at (4,1.5) 
	{Symbolic Domain \Z};
\node[inner sep=5pt] (bv_rep) at (4,1) 
{$\mathsf{100010001010100100110}$};

\draw[->,thick,bend right] ($(lies.north east)!0.5!(lies.south east)$) 
							to [out=320,in=150] 
							node [pos=0.55,yshift=-4ex,xshift=4ex,align=center,text width=3cm] 
								{\small Grounding \\ Function $f$}
							(bv_rep.west);

\node[inner sep=1pt] (logic) at (7.5,-1.5) 
	{$
   \begin{array}{ll}
    &
    |\{a : \mathsf{guilty}(a) \}| = 1
    \\
    \land &
    \forall a. |\{s : \mathsf{speaker}(a, s) \}| = 1
    \\
    \land &
    \forall s. |\{a : \mathsf{subject}(a, s) \}| = 1
    \\ 
    \land & 
    \forall s, a_1, a_2. \mathsf{speaker}(a_1, s) \land \mathsf{subject}(a_2, s) \limply \\ &\ \ \ \ \textbf{}
    	(\lnot\mathsf{guilty}(a_1) \lequiv \mathsf{accuses}(s) \lequiv \mathsf{guilty}(a_2))
\end{array}    
$};

\node[inner sep=5pt,text width=3cm,align=center,below of=logic,yshift=-2ex] (f) {Prediction Logic $\phi$};

\node[inner sep=0pt,text width=3cm,align=center] (f) at (7.5,1.5) 
	{Labels (Liar) \Y};
\node[inner sep=5pt] (result) at (7.5,1) 
	{$\{\mathsf{2}\}$};

\draw[->,thick] (bv_rep.south) to [out=270,in=90] ([xshift=-13ex]logic.north);
\draw[->,thick] ([xshift=-3.5ex]logic.north) to [out=100,in=270] 
				node [pos=0.5,xshift=7ex,align=center,text width=3cm] {\small Satisfying \\ Assignments} 
				(result.south);
\end{tikzpicture}
\end{center}
\caption{\label{fig:liarspuzzle} Liar's Puzzle example.}
\end{figure*}
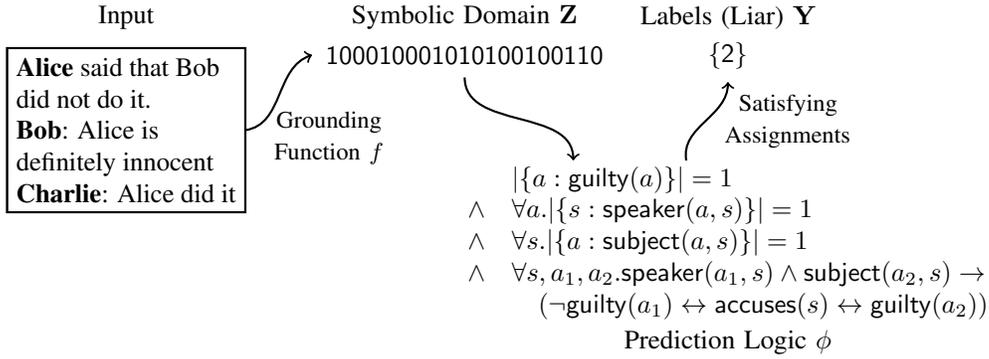

\subsection{Hyperparameters}
\label{sect:hypers}

\begin{table*}
\small
\centering
\begin{tabular}{l|cccccccc}
\toprule
\toprule
 & \multicolumn{2}{c|}{\emph{Conventional}} & \multicolumn{2}{|c}{\emph{w/} \layername} & \multicolumn{2}{|c}{\emph{w/} SATNet} & \multicolumn{2}{|c}{\emph{w/} Scallop}\\
 & \emph{optimizer} & \multicolumn{1}{c|}{\emph{epochs}}  & \emph{optimizer} & \multicolumn{1}{c|}{\emph{epochs}}  & \emph{optimizer} & \multicolumn{1}{c|}{\emph{epochs}}  & \emph{optimizer} & \emph{epochs} \\ \midrule
MNIST+ 10\% & SGD(1.0) & 0/5 &  SGD(1.0) & 3/5 & Adam(2e-3, 1e-5) & 3/5 & Adam(1e-3) & 3/5 \\
MNIST+ 25\% & SGD(1.0) & 0/5 & SGD(1.0) & 3/5 & Adam(2e-3, 1e-5) & 3/5 & Adam(1e-3) & 3/5 \\
MNIST+ 50\% & SGD(1.0) & 0/5 &  SGD(1.0) & 3/5 & Adam(2e-3, 1e-5) & 3/5 & Adam(1e-3) & 3/5 \\
MNIST+ 75\% & SGD(1.0) & 0/5 & SGD(1.0) & 3/5 & Adam(2e-3, 1e-5) & 3/5 & Adam(1e-3) & 3/5 \\
MNIST+ 100\% & SGD(1.0) & 0/5 & SGD(1.0) & 3/5 & Adam(2e-3, 1e-5) & 3/5 & Adam(1e-3) & 3/5 \\
Vis. Alg. \#1 & SGD(1.0) & 0/5 & SGD(1.0) & 3/5 & Adam(2e-3, 1e-5) & 3/5 & Adam(1e-3) & 3/5 \\
Vis. Alg. \#2 & SGD(1.0) & 0/5 & SGD(1.0) & 3/5 & Adam(2e-3, 1e-5) & 3/5 & Adam(1e-3) & 3/5 \\
Liar's Puzzle & Adam(2e-3) & 0/15 & Adam(1e-3) & 15/5 & Adam(2e-3, 1e-5) & 15/5 & --- & --- \\
Vis. Sudoku 10\% & SGD(1.0) & 0/100 & SGD(1.0) & 30/15 & Adam(2e-3, 1e-5) & 30/100 & --- & ---\\
Vis. Sudoku 50\% & SGD(1.0) & 0/100 & SGD(1.0) & 30/5 & Adam(2e-3, 1e-5) & 30/100 & --- & --- \\
Vis. Sudoku 100\% & SGD(1.0) & 0/100 & SGD(1.0) & 30/5 & Adam(2e-3, 1e-5) & 0/100 & --- & --- \\
\bottomrule
\end{tabular}

\caption{\label{tab:hyperparameters} Training hyperparameters. Numbers in parentheses after the optimizer denote the learning rate; when there are multiple numbers, different learning rates were applied to different parameter groups, as detailed in the text. Each epoch pair corresponds to the pre-training and training phases, i.e., 3/5 denotes three epochs of pre-training and five subsequent training epochs with the solver layer. }
\end{table*}

The four problems that our evaluation studies vary considerably in size and complexity, and the models used to train them require different considerations.
This section details these differences.
Table~\ref{tab:hyperparameters} relates the optimizers and epochs for each dataset and solver layer configuration.
For SATNet and Scallop, we use the optimization settings described in their respective papers, and found in their public implementations.
For SATNet models, the MaxSAT clause parameters were trained at a rate of 2e-3, and the convnet at rate 1e-5.
When SGD(1.0) is stated, we used a warmup period spanning the first epoch, and cosine annealing for the remainder of training.

\paragraph{MNIST Addition.}
The MNIST addition problem is the easiest of the problems that we study, at least in its full (100\%) configuration.
We find that for the 100\% configuration, all of the solver layers converge to a nearly optimal solution with three epochs of supervised pre-training, and five epochs of subsequent training with the solver layer attached.
For the subsample configurations, all of the solver layers converge to a stable, although in many cases suboptimal, solution within these parameters as well.
For \layername, we clipped gradients for all parameters at 0.1, and did not clip gradients for the other solver layer models.
For all configurations, we used batches of size 128.

\paragraph{Visual Algebra.}
Although visual algebra is a more difficult learning problem than MNIST addition, as evidenced by the  results in Table~\ref{tab:results}, we find that the same parameters allow all of the configurations studied in our evaluation to converge.
After 3/5 epochs of training, the models stabilize, and in some cases, further training yields an overfit model.
For \layername, we clipped gradients for all parameters at 0.1, and did not clip gradients for the other solver layer models.
For all configurations, we used batches of size 64.

\paragraph{Liar's Puzzle.}
The Liar's Puzzle is the only problem to use a recurrent model, and we found that it required more epochs of pre-training to reduce the variance of the final model with the solver layer.
Additionally, the SGD optimizer used with \layername on other datasets caused the model to converge at local minima.
We found that pre-training at a higher learning rate, and using Adam with a default learning rate, let to the best results.
Additionally, we did not clip gradients for the \layername model.
We used the same parameters, but the normal SATNet optimizer, for the SATNet model.
For all configurations, we used batches of size 32.

\paragraph{Visual Sudoku.}
Visual Sudoku is the most challenging problem that we studied, for all solver layers as well as the conventional model.
We did not use supervised pre-training, as the supervision in this problem leaks the correct labels directly to the model, bypassing the solver layer and the need for its updates~\cite{chang2020assessing}.
We instead used the unsupervised pre-training method described in~\cite{topan2021techniques}, and found that 30 epochs of unsupervised pre-training was sufficient to yield consistent and quick convergence with the solver layer.
For the most data-scarce configuration (10\%), 15 epochs of training with the solver layer were needed to converge, and for the others five epochs were sufficient.
For the \layername model, we used batches of size 1 after pretraining (batch size 64 during pre-training), primarily due to the fact that \layername returns only the indices masked as non-hint elements on the Sudoku board.
Because each instance has a different number of hint elements, this would lead to ragged tensors during training, which Pytorch does not support.

When assessing SATNet on Visual Sudoku, we were unable to converge to a useful model on any except the full (100\%) configuration, as discussed in Section~\ref{sect:eval-results}.
Using the authors' public implementation and training script, we attempted the 10\% and 50\% configurations with and without unsupervised pre-training, with and without the measures taken in~\cite{chang2020assessing} to prevent label leakage, and with batches of size 40 (as used in the original paper) as well as 1, to no avail.
For the 100\% configuration, we were able to reproduce a useful model; we use the accuracy reported in the original paper in Table~\ref{tab:results} for consistency, as the average that we obtained did not differ significantly from this.


\end{document}